\pdfoutput=1

\documentclass[runningheads]{llncs}

\usepackage[T1]{fontenc}

\usepackage{graphicx}
\usepackage{comment}
\usepackage{amsmath,amssymb}
\usepackage{color}
\usepackage{url}

\newif\ifreview
\reviewfalse

\ifreview
	\usepackage{lineno}

	\linenumbers
\fi

\usepackage{annotate-equations}

\def\eg{\emph{e.g.}} 
\def\ie{\emph{i.e.}} 
\def\etal{\emph{et~al.}} 

\begin{document}


\def\SubNumber{007}

\def\GCPRTrack{Main Track}

\title{HiFiHR: Enhancing 3D Hand Reconstruction from a Single Image via High-Fidelity Texture}

\ifreview
\else
        \titlerunning{Enhancing 3D Hand Recon. from a Single Image via High-Fidelity Texture}

         \author{Jiayin Zhu \and
            Zhuoran Zhao \and
            Linlin Yang \and
            Angela Yao }
	
	\authorrunning{J. Zhu, Z. Zhao, et al.}
	
       \institute{National University of Singapore \\
        \email{\{zhujiayin,zhuoran.zhao\}@u.nus.edu, \{yangll,ayao\}@comp.nus.edu.sg}
            }
\fi

\maketitle              

\begin{abstract}
We present HiFiHR, a high-fidelity hand reconstruction approach that utilizes render-and-compare in the learning-based framework from a single image, capable of generating visually plausible and accurate 3D hand meshes while recovering realistic textures. Our method achieves superior texture reconstruction by employing a parametric hand model with predefined texture assets, and by establishing a texture reconstruction consistency between the rendered and input images during training. Moreover, based on pretraining the network on an annotated dataset, we apply varying degrees of supervision using our pipeline, \ie, self-supervision, weak supervision, and full supervision, and discuss the various levels of contributions of the learned high-fidelity textures in enhancing hand pose and shape estimation. Experimental results on public benchmarks including FreiHAND and HO-3D demonstrate that our method outperforms the state-of-the-art hand reconstruction methods in texture reconstruction quality while maintaining comparable accuracy in pose and shape estimation. Our code is available at \url{https://github.com/viridityzhu/HiFiHR}.

\keywords{3D Hand Reconstruction  \and 3D from Single Images.}
\end{abstract}
\section{Introduction}
    With the development of VR/AR, photo-realistic 3D reconstruction has gained raising attention, and significant progress has been made in the human body and face~\cite{jiang2022neuman,xu2021hnerf,xu2022surface,su2021anerf,hong2021headnerf}.
    Hands, as the main medium through which people interact with the world, are also essential to be accurately reconstructed.
    
    Existing works~\cite{chen2021s2hand,boukhayma20193d,Moon_2020_ECCV_DeepHandMesh,HTML_eccv2020,karunratanakul2022harp} leverage differentiable rendering~\cite{chen2019dibr} and a parametric hand model, such as MANO~\cite{mano}, to reconstruct 3D hand meshes from images. These methods excel in achieving precise pose estimation while also demonstrating high efficiency during inference.
    Another recent line of work adapts NeRF-based~\cite{su2021anerf} 3D reconstruction by modeling geometry and texture properties together from ray queries, resulting in high-quality textures~\cite{corona2022lisa,chen2022handavatar,mundra2023livehand}.
    However, these methods have limitations: (1) Render-based approaches often neglect texture reconstruction or achieve limited accuracy using simplistic representations.
    (2) NeRF-based methods require videos or multi-view images as input, and suffer from high computational complexity and limited generalization.
    
    In practical applications, obtaining multi-view images can be time-consuming or unfeasible, leaving us with scenarios where only a single image is available. In such cases, render-based methods remain the most viable option. However, monocular images of hands commonly suffer from severe occlusion and depth ambiguity, posing challenges for reconstructing plausible hand structures. This motivates the need for high-fidelity texture reconstruction, as it has the potential to facilitate  accurate estimation of hand pose and shape. 
    In contrast, NIMBLE~\cite{nimble}, an anatomy-based hand model, embeds numerous texture assets and leverages physical-based rendering~\cite{PHARR20171pbr} for a high-fidelity texture representation. Moreover, it physically constrains the texture by the relative motion between muscles, bones, and skins, which leads to reliable estimation results. 
    Leveraging this novel texture representation holds great potential for enhancing hand reconstruction quality.
    
\begin{figure}
    \includegraphics[width=\textwidth]{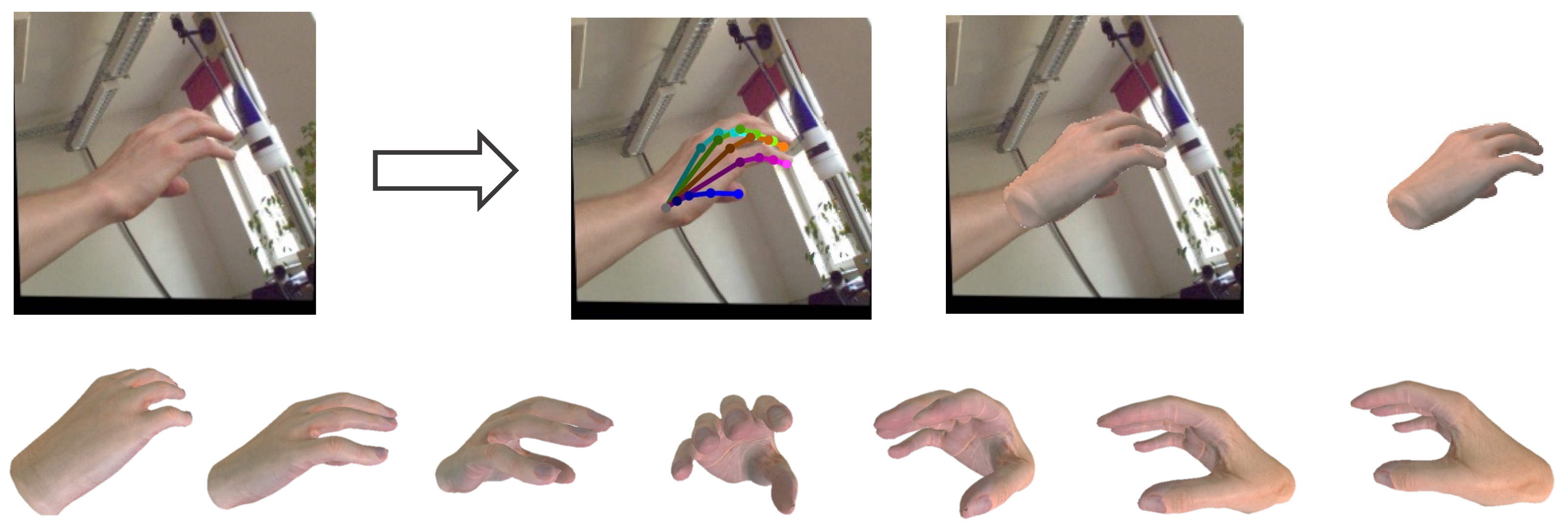}
    \caption{Our method reconstructs a realistic 3D hand from a monocular image leveraging the NIMBLE \cite{nimble} hand model and the proposed texture reconstruction consistency. It synthesizes lifelike hands from just a single view, delivering plausible representations from every perspective. }
    \label{fig1}
    \end{figure}

    Building upon these challenges, our work focuses on achieving high-fidelity texture reconstruction in single-image scenarios. 
    Leveraging the advantage of NIMBLE~\cite{nimble} in a rendering pipeline, we build a HiFiHR (High-fidelity hand reconstruction) model which is able to predict 3D hand pose, shape, texture, and lighting from single input images (Fig.~\ref{fig1}).

    Furthermore, previous works have attempted to enhance pose estimation through texture reconstruction. However, these efforts have primarily concentrated on single supervision settings, such as self-supervision by S$^2$HAND~\cite{chen2021s2hand} and weak supervision by SMHR~\cite{Ren2023EndtoEndWS}. Additionally, the limited quality of the reconstructed textures in these approaches hinders their performance. As a result, no significant conclusions have been drawn, leaving room for further exploration. In our work, we comprehensively investigate this question across various levels of supervision, aiming to ascertain the extent to which high-fidelity texture reconstruction consistency can aid the learning of pose and shape.

    The main contributions of this work can be concluded as follows.
    \begin{enumerate}
        \item By leveraging the advantages of model-based methods, our approach produces high-fidelity and consistent hand textures from a single input image, resulting in a more realistic representation of the reconstructed 3D hands.
        \item We investigate the impact of hand texture reconstruction on pose and shape estimation under varying levels of supervision. Our findings reveal that high-fidelity texture consistency aids pose and shape learning in self-supervision, but introduces noise in weak supervision. And the effect is minimal in full supervision with stronger constraints from 3D labels.
        \item Quantitative and qualitative experiments on two single-hand reconstruction benchmarks, \ie, FreiHAND and HO-3D, verify the effectiveness of our approach in both pose and shape accuracy and texture reconstruction quality.
    \end{enumerate}

\section{Related Work}

\subsection{Parametric Hand Models}
Given parameters on hand pose and shape, parametric hand models infer a hand mesh with vertices and faces. MANO~\cite{mano} is a widely used hand model, which follows SMPL~\cite{SMPL_2015} to adopt linear blend skinning to deform a mesh based on a kinematic skeleton. However, MANO does not capture a texture space.
One follow-up work, HTML~\cite{HTML_eccv2020}, captures hand texture from two image sequences and fits a MANO template to extract the texture from the scanned mesh. Furthermore, an anatomy-based hand model, NIMBLE~\cite{nimble}, creates a better hand texture model via PCA and pre-defined texture assets. The photorealistic texture is represented with diffuse, normal, and specular maps, and a more reliable hand pose can be achieved by enforcing inner bones and muscles to match anatomic and kinematic rules. 

\subsection{Hand Texture Reconstruction}
Boukhayma \etal~\cite{boukhayma20193d} introduce the first end-to-end deep learning-based method that predicts both 3D hand shape and pose from RGB images in the wild. It leverages MANO as a pre-computed hand model and a re-projection module. 
Similarly, S$^2$HAND~\cite{chen2021s2hand} and SMHR~\cite{Ren2023EndtoEndWS} achieve 3D hand reconstruction with texture, also adopting MANO as a pre-computed hand model. However, these works achieve limited texture reconstruction because of the simple texture representation, \ie, RGB values of each vertex in the mesh.
DeepHandMesh~\cite{Moon_2020_ECCV_DeepHandMesh} mainly focuses on shape reconstruction, while also providing a method to unwrap multiview RGB images to a high-quality $1024\times 1024$ texture map. Moreover, 
LiveHand~\cite{mundra2023livehand} and HandAvatar~\cite{chen2022handavatar} are two recent NeRF-based works that also achieve photo-realistic texture, but they are not suitable in scenarios when there is only one image available. 

Differently, this work aims to achieve high-fidelity 3D hand texture reconstruction from a single image. Based on NIMBLE, a 3D hand mesh with high-resolution texture UV maps can be estimated from the 2D image. By employing a differentiable renderer, weak supervision on hand texture is achieved through consistency between the rendered 2D image and the input image. Additionally, the pipeline incorporates pre-computed texture assets in NIMBLE, enabling the generation of plausible texture estimations for unseen or occluded parts of the hand, even when the input is derived from a monocular view. The iterative render-and-compare loop further ensures accurate alignment of the reconstructed 3D hand with the 2D image.

\section{Methods}

\subsection{Overview}

    \begin{figure}
    \includegraphics[width=\textwidth]{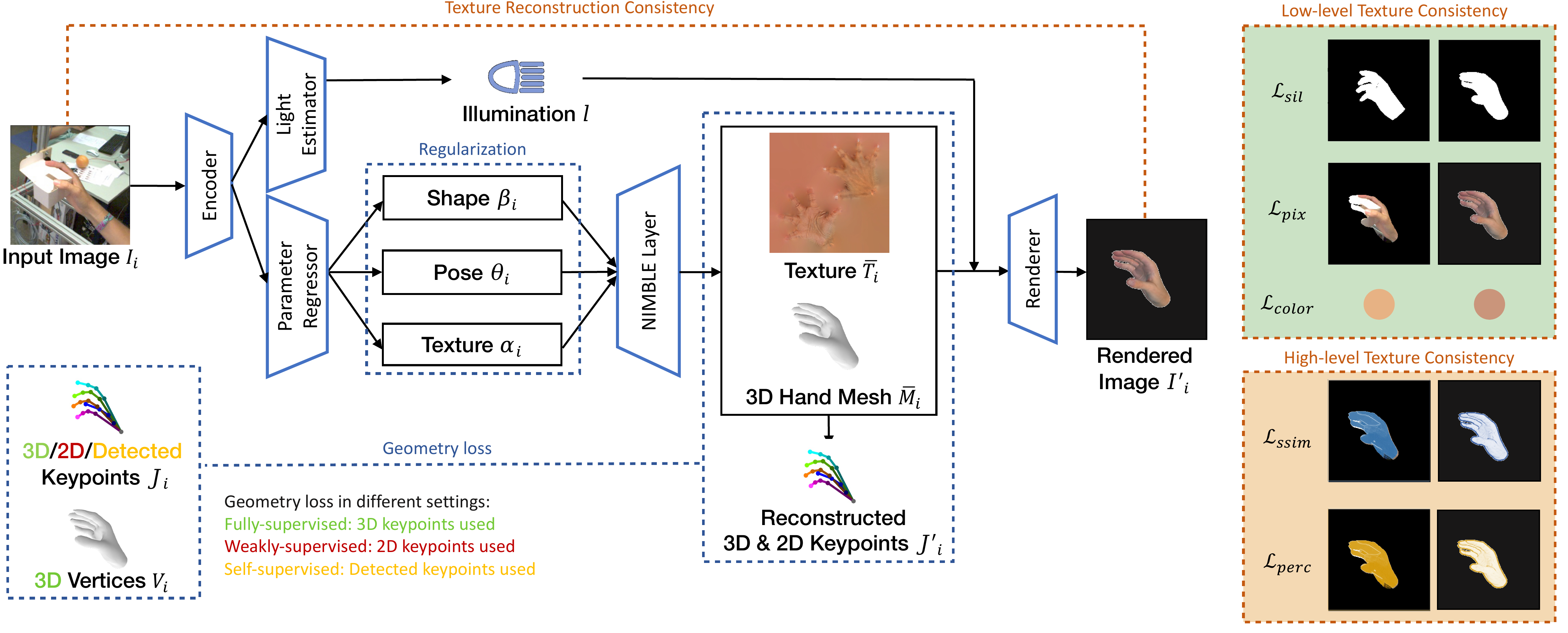}
    \caption{Overview of our 3D hand reconstruction pipeline. Given one single-view image, our method generates a plausible and high-fidelity 3D textured hand mesh. The method is supervised by geometry loss, texture reconstruction consistency loss, and regularization. Geometry loss adopts different settings under different kinds of supervision. On the right are details of the texture reconstruction consistency, which consists of a low-level consistency and a high-level consistency.} \label{fig:overview}
    \end{figure}

    \noindent Our end-to-end pipeline is able to reconstruct 3D hand given a single RGB image, as demonstrated in Fig.~\ref{fig:overview}.
    Given the input image $I_i$, where $i\in [1,N]$ and $N$ is the number of samples in the dataset, we adopt an encoder $\mathcal E$, a NIMBLE~\cite{nimble} hand layer $\mathcal N$, and a differentiable renderer $\mathcal R$, to obtain a textured 3D hand mesh $M_i$ and a rendered image $I_i'$.

\subsection{Model Structure}

    \subsubsection{Parameter Encoder.}
    An encoder $\mathcal E$ is trained to estimate the parameters including shape $\beta_i$, pose $\theta_i$, texture $\alpha_i$, and illumination $l_i$:
    \begin{equation}
        \{\beta_i, \theta_i, \alpha_i, l_i\} = \mathcal{E}(I_i).
    \end{equation}
    The encoder consists of a visual feature extractor, \eg,  Efficientnet~\cite{tan2019efficientnet}, and independent MLP layers for each parameter. $\beta_i$, $\theta_i$, and $\alpha_i$ are NIMBLE parameters controlling hand shape, pose, and appearance. They are vectors of length $20$, $30$, and $10$, respectively, following the default setting of NIMBLE.

    Disentangling lighting from texture is a crucial factor for obtaining plausible hand texture. This is because the feasible texture space of NIMBLE is limited to natural skin tones, while images in the wild often have more complex lighting conditions such as colorful lights that NIMBLE cannot fit. 
    Therefore, we also estimate illumination $l_i$ from the input image, which is a vector of length $6$, defining directional lighting from a direction $(x_i,y_i,z_i)$ with color $(r_i,g_i,b_i)$:
    \begin{equation}
        l_i = (x_i,y_i,z_i,r_i,g_i,b_i).
    \end{equation}
    \subsubsection{NIMBLE Layer.}
    NIMBLE takes the parameters estimated by the encoder $\mathcal E$ as input and generates a realistic 3D hand with bone, muscle, and skin geometry, as well as a photo-realistic appearance:
    \begin{equation}
       \{\bar M_i, \bar T_i\} = \mathcal{N}(\beta_i, \theta_i, \alpha_i),
    \end{equation}
    where $\bar M_i = \{M_i^{skin}, M_i^{bone}, M_i^{muscle}\}$, which consists of a skin mesh, a bone mesh, and a muscle mesh. And $\bar T_i = \{T_i^{diff}, T_i^{spec}, T_i^{norm}\}$, which consists of a diffuse map, a specular map, and a normal map. 

    The skin mesh is physically conditioned by the bone mesh and muscle mesh to obtain plausible hand skin, which contains $5990$ vertices and $9984$ faces. $25$ joints are obtained using a regressor from the skin mesh. However, commonly used datasets only provide annotations for the $778$ vertices and $21$ joints defined by MANO, and the $25$ NIMBLE joints are not compatible. Thus, an additional linear layer is utilized to regress the $778$ MANO vertices from the skin mesh, denoted as $M_i$. Additionally, we adopt the MANO regressor to obtain $21$ MANO joints, denoted as $J_i^{3D}$, from $M_i$.

    \subsubsection{Differentiable Render.}
    The generated textured mesh is then rendered back into the image domain as $I_i'$ by a differentiable neural renderer PyTorch3D~\cite{ravi2020pytorch3d} utilizing the illumination $l_i$ \cite{chen2019dibr}:
    \begin{equation}
        I_i' = \mathcal{R}(M_i, T_i^{diff}, l_i, k_i),
    \end{equation}
    where $k_i$ is the ground truth camera intrinsic parameters.
    The 2D keypoints $J_i'$ are also re-projected from 3D joints $J_i^{3D}$ given $k_i$.
    Note that because the PyTorch3D renderer~\cite{ravi2020pytorch3d} does not support physical-based rendering yet, we only use $T_i^{diff}$ as the texture UV mapping. 

\subsection{Training Objective}
    We train the pipeline using single-view hand images. The optimization objective is to minimize the difference between the input and the rendered hand images, with different levels of supervision on joints and vertices.
    The overall training loss $\mathcal L$ consists of three parts including geometry loss $\mathcal L_{geo}$, texture reconstruction consistency loss $\mathcal{L}_{tex}$, and regularization $\mathcal L_{regu}$:
    \begin{equation}
        \mathcal L = w_{geo}\mathcal L_{geo} + w_{tex}\mathcal{L}_{tex} + w_{regu}\mathcal L_{regu},
    \end{equation}
    where $w_{geo}$, $w_{tex}$, and $w_{regu}$ are weighting factors for each loss term.
    
    \subsubsection{Geometry Loss.} 
    The geometry of the mesh is constrained with the following terms:
    \begin{equation}
        \mathcal L_{geo}^{3D} = w_{jnt}\mathcal  L_{jnt} +  w_{vert}\mathcal  L_{vert} +  w_{direc}\mathcal  L_{direc} +  w_{len}\mathcal  L_{len},
    \end{equation}
    \begin{equation}
        \mathcal L_{geo}^{2D} = w_{jnt}\mathcal  L_{jnt}^{2D}  +  w_{direc}\mathcal  L_{direc}^{2D}.
    \end{equation}
    $w_{jnt}$, $w_{vert}$, $w_{direc}$, and $w_{len}$ are weighting factors for each loss term.
    In the full 3D supervision scheme, we train the pipeline using $\mathcal L_{geo}^{3D}$ where all the loss terms are in 3D space, \eg, the 3D joint loss $L_{jnt}$. While in the weak 2D supervision scheme, we only employ $\mathcal L_{geo}^{2D}$ where all the loss terms are in 2D space after projection. 
    $\mathcal L_{jnt}$ and $\mathcal L_{vert}$ penalize the $l_1$-distance between the predicted and ground-truth joint and vertex positions, respectively. $\mathcal L_{direc}$ is the loss on bone directions, where a bone is defined as the vector between two adjacent joints. $\mathcal L_{len}$ penalizes the difference in edge lengths of all faces. 
    Furthermore, in the self-supervision scheme, we only utilize  $\mathcal L_{jnt}^{2D}$  and $\mathcal L_{direc}^{2D}$  with detected 2D joints obtained through OpenPose~\cite{cao2017realtime} at an offline stage, along with corresponding bone directions. 
    The confidence score provided by OpenPose is integrated into  $\mathcal L_{jnt}^{2D}$ and $\mathcal L_{direc}^{2D}$, which shows the probability of joints being detected correctly. Please refer to the supplementary for more information on the losses.
    
    \subsubsection{Texture Reconstruction Consistency.}
    The texture reconstruction consistency is designed to ensure the consistency between the input and rendered images, and is formulated as:
    \begin{equation}
        \mathcal L_{tex} = \eqnmarkbox[blue]{node1}{w_{pix}\mathcal  L_{pix} + w_{color}\mathcal  L_{color} + w_{sil}\mathcal  L_{sil}} + \eqnmarkbox[red]{node2}{w_{ssim}\mathcal  L_{ssim} + w_{perc}\mathcal  L_{perc}}.
    \end{equation}
    \annotate[yshift=-0.5em]{below,left}{node1}{$\mathcal L_{low\text{-}level}$}
    \annotate[yshift=-0.5em]{below}{node2}{$\mathcal L_{high\text{-}level}$}
    \vspace{1em}

    $\mathcal L_{pix}$ is a per-pixel $l_1$ loss between the input and rendered images, $\mathcal L_{color}$ encourages the mean RGB colors to be close, $\mathcal L_{sil}$ penalizes the difference between silhouettes, $\mathcal L_{ssim}$ measures the structural similarity between the two images~\cite{ssim}, and $\mathcal L_{perc}$ is the LPIPS loss~\cite{ledig2017photo,zhang2018lpips} that captures the perceptual difference between the two images by comparing the features extracted using the AlexNet model~\cite{krizhevsky2017imagenet}. We use a hand mask to extract the foreground hand of the input image to make a comparison with the rendered hand. In the full supervision and weak supervision schemes, we use the ground-truth hand mask. In the self-supervision scheme, we use the predicted hand mask. $w_{pix}$, $w_{color}$, $w_{sil}$, $w_{ssim}$, and $w_{perc}$ are weighting factors for each loss term.

    Among these texture reconstruction consistency terms, the low-level texture consistency $\mathcal L_{low\text{-}level}$ consists of $\mathcal  L_{pix}$, $\mathcal  L_{color}$, and $\mathcal  L_{sil}$, which operates at the low level of pixel values and results in sensitivity to geometry misalignment. To mitigate this issue, we also include the high-level texture consistency $\mathcal L_{high\text{-}level}$, consisting of $ \mathcal  L_{ssim}$ and $\mathcal  L_{perc}$, to enable abstract comparisons of the textures and align more closely with human perceptual results.

    \subsubsection{Regularization.}
    We also add $l_2$-regularizers on the magnitude of the shape, pose, and texture parameters to ensure the results are plausible, where $w_{\beta}$, $w_{\theta}$, and $w_{\alpha}$ are weighting factors for each loss term:
    \begin{equation}
        \mathcal L_{regu} = w_{\beta}\|\beta\|_2 + w_{\theta}\|\theta\|_2 + w_{\alpha}\|\alpha\|_2.
    \end{equation}

\section{Experiment}

\subsection{Implementation Details}
The proposed method is implemented with PyTorch~\cite{paszke2017automatic}. We use EfficientNet \cite{tan2019efficientnet} pretrained on the ImageNet dataset as our feature extractor. The input to our model is a 224 $\times$ 224 RGB image. In our training process, the batch size is set to 50. The initial learning rate is $10^{-3}$ and is decreased by 2 at the epoch of 50, 80, 110, and 160. We use Adam \cite{kingma2014adam} for optimization. We train our model on one NVIDIA QUADRO RTX 8000 GPU. In the full supervision setting, we mainly refer to existing works~\cite{chen2021s2hand,Ren2023EndtoEndWS} to set $w_{geo}=100, w_{tex}=0.01, w_{regu}=0.01, w_{jnt}=2, w_{vert}=1.5, w_{direc}=0.06, w_{len}=1, w_{pix}=2, w_{color}=0.2, w_{sil}=0.08, w_{ssim}=1, w_{perc}=10^{-6}$. More discussion of weight ablation studies can be seen in the supplementary materials.

\subsection{Datasets and Evaluation Metrics}

Our pipeline is initially pretrained on a synthetic dataset, either RHD or DARTset, and then train and evaluate on two real-world datasets, including FreiHAND and HO-3Dv2.

\noindent\textbf{RHD}~\cite{zb2017RHD}. RHD is a considerable synthetic dataset that includes 3D annotations for both singular and interacting hand postures. The dataset synthesizes hand positions from an array of $20$ unique characters engaged in a diverse set of $39$ actions. 
It offers a substantial supply of images for empirical study, with $41,258$ training instances and $2,728$ instances for evaluation, thereby providing a slight perturbation in the data distribution.

\noindent\textbf{DARTset}~\cite{gao2022dart}. DARTset is a large-scale hand dataset with diverse poses, textures, and accessories, comprising 800K samples split into a training set (758,378) and a test set (28,877). Among these hands, 25\% are assigned an accessory. 
In our experiment, we used 100,000 samples from the training set and 288,77 samples from the test set to pretrain our pipeline.

\noindent\textbf{FreiHAND}~\cite{zimmermann2019freihand}. FreiHAND is a commonly used dataset that contains single-hand images in the wild. There are $32,560$ images for training and $3,960$ images for testing. For each sample, one RGB image and annotations for 3D joints and vertices labels are provided.

\noindent\textbf{HO-3Dv2}~\cite{hampali2020ho3d}. HO-3D is 3D pose annotations for hands and objects under severe occlusions from each other.
It contains annotations for 77,558 images which are split into 66,034 training images (from 55 sequences) and 11,524 evaluation images (from 13 sequences).
Evaluations are conducted by submitting our estimated results to their online system.

\noindent\textbf{Evaluation Metrics.}
As with existing works~\cite{chen2021s2hand,boukhayma20193d}, we use the mean per joint position error (MPJPE) and mean per vertex position error (MPVPE) in cm between the prediction and ground truth after Procrustes alignment to evaluate the 3D reconstruction accuracy of the joints and mesh vertices. 
To evaluate the texture reconstruction accuracy, we use metrics that focus on the rendered image quality. The L1 distance of two images is used for a low-level representation of the reconstruction quality. SSIM~\cite{ssim} and PSNR are used to reflect image similarity as the rendering quality. Besides, consistent with some recent works \cite{weng_humannerf_2022_cvpr,chen2022handavatar}, we also adopt LPIPS~\cite{zhang2018lpips} as high-level metrics representing the human perception of the texture quality.

\begin{table}
\begin{center}
\caption{Comparison on FreiHAND~\cite{zimmermann2019freihand}. The column ``Supv.'' indicates whether the supervision level is 2D or 3D. The column ``Tex.'' indicates whether the method is able to reconstruct hand texture.}\label{tab_freihand}
\begin{tabular}{|l|c|cc|c|cccc|}
\hline
Methods & Supv. & MPJPE$\downarrow$ & MPVPE$\downarrow$& Tex. & L1$\downarrow$ & PSNR$\uparrow$ & SSIM$\uparrow$ & LPIPS$\downarrow$ \\
\hline
Biomechanical~\cite{spurr2020weakly}& 2D & 1.13 & - & No & - & - & - & - \\
S$^2$HAND~\cite{chen2021s2hand}& 2D & 1.18 & 1.19 & Yes & 0.12 & 16.61 & 0.79 & 0.43\\ 
SMHR~\cite{Ren2023EndtoEndWS}& 2D & 1.07 & 1.10 & Yes & - & 16.64 & - & - \\

Ours$_\text{weak}$  & 2D   & 1.23 & 1.24 & Yes & 0.02 & 20.00 &  0.93 &  \textbf{0.09} \\
Ours$_\text{self}$  & 2D   & 1.31 & 1.33 & Yes & \textbf{0.02} & \textbf{20.04} & \textbf{0.94} & 0.10 \\
Ours$_\text{self*}$  & 2D   & 1.66 & 1.65 & Yes & 0.03 & 17.88 & 0.92 & 0.13 \\
\hline
Boukhayma \etal~\cite{boukhayma20193d} & 3D & 3.50 & 1.32 & No & - & - & -  & - \\
ObMan~\cite{hasson2019learning} & 3D & 1.33 & 1.33  & No & - & - & - & - \\
ManoCNN~\cite{zimmermann2019freihand} & 3D & 1.10 & 1.09   & No & - & - & - & -  \\
ManoFit~\cite{zimmermann2019freihand} & 3D & 1.37 & 1.37   & No & - & - & - & -  \\
HTML~\cite{HTML_eccv2020}$^\dagger$ & 3D & 1.11 & 1.10 & Yes & - & - & - & -  \\
HIU~\cite{zhang2021hiu} & 3D & 0.71 & 0.86 & No & - & - & - & -  \\
SMHR~\cite{Ren2023EndtoEndWS}& 3D & 0.80 & 0.81 & Yes & - & 16.64 & - & - \\

Ours$_\text{full}$  & 3D   & 1.21 & 1.23 & Yes & 0.03 &  \textbf{19.55} & 0.94 & 0.10 \\
\hline
\end{tabular}
\end{center}\vspace{-.1in}
\footnotesize{$^\dagger$: Due to the unavailability of HTML's 3D reconstruction implementation codes, which were not published alongside their hand layer codes, we cannot reproduce their results or provide texture reconstruction metrics for comparison.
}
\end{table}

\subsection{Comparison to State-of-the-art Methods}
In this section, we evaluate the reconstruction performance of our approach and compare it with the state-of-the-art methods on two widely used single-hand datasets, FreiHAND~\cite{zimmermann2019freihand} and HO-3Dv2~\cite{hampali2020ho3d}. Following SMHR~\cite{Ren2023EndtoEndWS}, we mainly focus on the model-based methods for a fair comparison.

\noindent\textbf{Comparison on Geometry Reconstruction Quality.}
We first report the comparison results on the geometry reconstruction using MPJPE and MPVPE with state-of-the-art methods~\cite{spurr2020weakly,chen2021s2hand,Ren2023EndtoEndWS,boukhayma20193d,hasson2019learning,zimmermann2019freihand,HTML_eccv2020,zhang2021hiu}.
Note that most existing works focus on the 3D geometry reconstruction and ignore textures, while our work is able to reconstruct the geometry as well as the high-quality texture. 

Tab.~\ref{tab_freihand} shows the results on FreiHAND~\cite{zimmermann2019freihand}, and Tab.~\ref{tab_ho3d} shows the results on HO-3Dv2~\cite{hampali2020ho3d}. 
Ours$_\text{full}$ is our method under full 3D supervision, \ie, supervised with ground truth 3D joints and vertices. Ours$_\text{weak}$ is weakly supervised by only 2D ground truth joints, \ie, $\mathcal L_{jnt}^{2D}$. 
While the definition of self-supervision is controversial, we implement two versions for discussion. Ours$_\text{self}$ follows S$^2$HAND~\cite{chen2021s2hand} to detect noisy 2D joints via OpenPose~\cite{openpose} and utilize the confidence-aware 2D joint loss, and another version, Ours$_\text{self*}$, does not adopt any form of 2D annotation and is only supervised by input images using texture reconstruction consistency losses.

\begin{table}
\centering
\caption{Comparison on HO-3Dv2~\cite{hampali2020ho3d}. The column ``Supv.'' indicates whether the supervision level is 2D or 3D. The column ``Tex.'' indicates whether the method is able to reconstruct hand texture.}\label{tab_ho3d}
\begin{tabular}{|l|c|cc|c|cccc|}
\hline
Methods & Supv. & MPJPE$\downarrow$ & MPVPE$\downarrow$& Tex. & L1$\downarrow$ & PSNR$\uparrow$ & SSIM$\uparrow$ & LPIPS$\downarrow$ \\
\hline
PeCLR~\cite{spurr2021peclr}& 2.5D & 1.09 & - & No & - & - & - & - \\
S$^2$HAND~\cite{chen2021s2hand}& 2D & 1.14 & 1.12 & Yes & - & 13.92 & - & - \\ 
SMHR~\cite{Ren2023EndtoEndWS}& 2D & 1.03 & 1.01 & Yes & - & 16.78 & - & - \\

Ours$_\text{weak}$  & 2D   & 1.16 & 1.23 & Yes & 0.01 & 23.66 & 0.95 & 0.12 \\
Ours$_\text{self}$  & 2D   & 1.23 & 1.26 & Yes & \textbf{0.01} & \textbf{26.50} & \textbf{0.96} & \textbf{0.10} \\
Ours$_\text{self*}$  & 2D   & 1.48 & 1.51 & Yes & - & - & - & - \\
\hline
HO3D~\cite{hampali2020ho3d} & 3D & 1.07 & 1.06 & No & - & - & -  & - \\
ObMan~\cite{hasson2019learning} & 3D & - & 1.10  & No & - & - & - & - \\
Photometric~\cite{hasson2020leveraging} & 3D & 1.11 & 1.14   & No & - & - & - & -  \\
SMHR~\cite{Ren2023EndtoEndWS}& 3D & 1.01 & 0.97 & Yes & - & 16.78 & - & - \\

Ours$_\text{full}$  & 3D   & 1.16 & 1.22 & Yes & {0.01} & \textbf{23.97} & {0.95} & {0.12} \\
\hline
\end{tabular}
\end{table}

Our method achieves comparable pose and shape accuracy with state-of-the-art methods under all kinds of supervision. 
Note that previous works like S$^2$HAND \cite{chen2021s2hand} and SMHR \cite{Ren2023EndtoEndWS} use additional branches to perform keypoint estimation or hand localization while our pipeline does not rely on an extra branch. Our focus lies in studying the impact of texture reconstruction on pose and shape learning across different levels of supervision, rather than solely aiming for the best results in pose and shape reconstruction. Consequently, we employ a simplified structure and training settings. Therefore, our experimental results remain competitive, demonstrating the robustness and effectiveness of our approach.

\noindent\textbf{Comparison on Texture Reconstruction Quality.}
Secondly, we compare our texture reconstruction quality with several existing methods that are able to generate hand textures~\cite{chen2021s2hand,Ren2023EndtoEndWS}. (See the texture metrics in Tab.~\ref{tab_freihand} and Tab.~\ref{tab_ho3d}.) 

\begin{figure}
    \includegraphics[width=\textwidth]{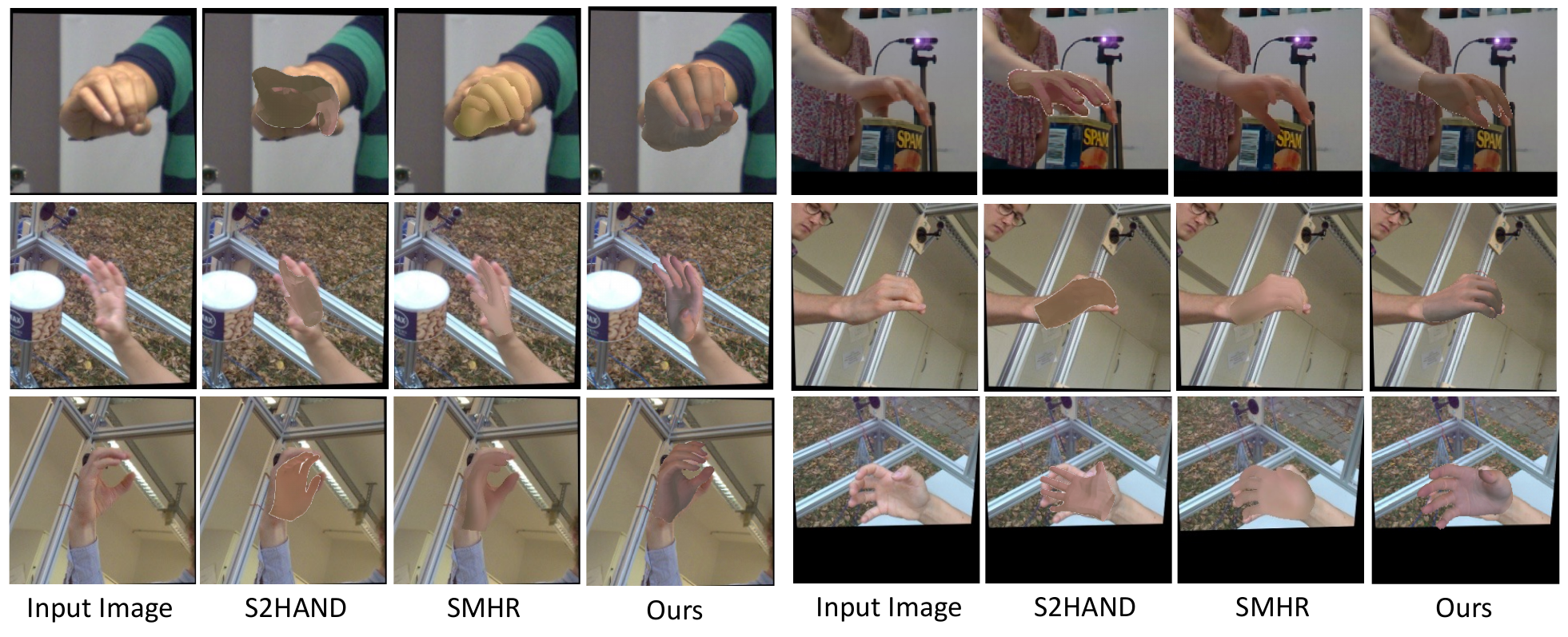}
    \caption{Visualization comparison of texture quality with the state-of-the-art methods on FreiHAND and HO-3Dv2. The results clearly demonstrate the superior accuracy, high fidelity, and overall plausibility of our reconstructed hand texture compared to other methods.} \label{fig_comparison}
\end{figure}

It can be clearly seen that our texture reconstruction quality outperforms all the methods. 
It achieves a $20.2\%$ increase on FreiHAND and a $57.9\%$ increase on HO-3Dv2 in PSNR compared to the previous state-of-the-art method, SMHR~\cite{Ren2023EndtoEndWS}. This is mainly because of our high-fidelity texture representation, \ie, physical-based rendering~\cite{PHARR20171pbr}, which represents texture using multiple high-resolution UV maps. Instead, S$^2$HAND~\cite{chen2021s2hand} and SMHR~\cite{Ren2023EndtoEndWS} simply represent textures as per-vertex RGB values, which are crude and lack realism.
Moreover, we achieve high LPIPS values, showing that our method produces high-fidelity hands in human perception.
Here we exclude the texture quality results of Self$^*$ from the HO-3Dv2 evaluation because of the significant influence of inaccurate pose and shape estimation on these results.

\begin{figure}
  
    \centering
    \includegraphics[width=\textwidth]{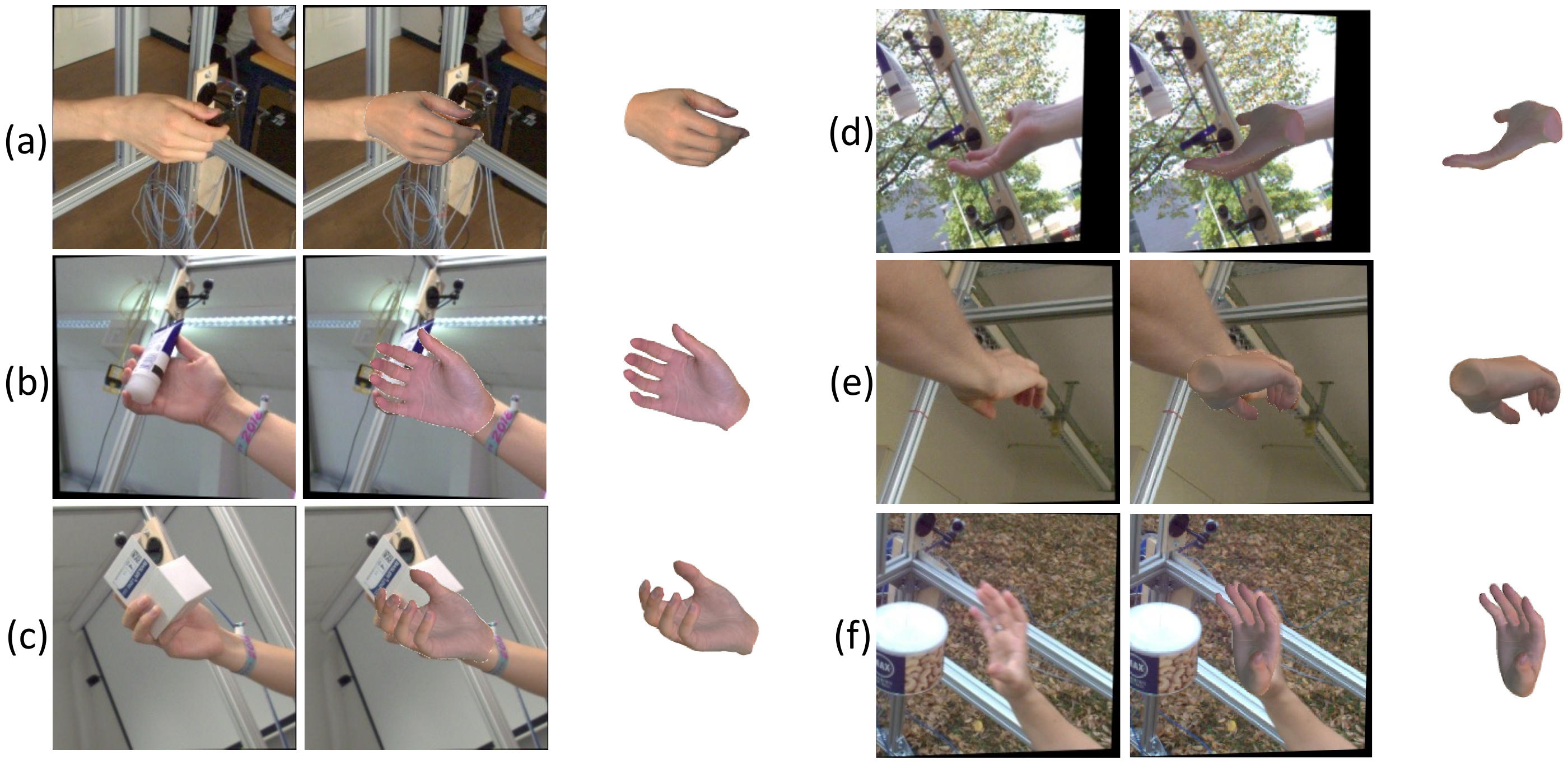}
    \caption{Qualitative hand reconstruction results on FreiHAND and HO-3Dv2. (a)(b) High-fidelity details such as skin bumps and palm prints. (c)(d)(e) Plausible hands even severely occluded, in extreme poses or viewpoints. (f) Plausible hand even when the input image is in motion blur.}
    \label{fig_qualitative}
\end{figure}

Fig.~\ref{fig_qualitative} showcases the visualization results of our reconstructed hands, highlighting the generation of high-fidelity hands with detailed skin features such as palm prints, nails, and bone bumps. Even in challenging scenarios with extreme poses and severe occlusions, our method produces plausible hand reconstructions despite limited information about the hand texture. Besides, Fig.~\ref{fig_comparison} visually compares our method with state-of-the-art approaches, demonstrating the superior accuracy, high fidelity, and plausibility of our reconstructed hand texture.

\subsection{Evaluation on Texture Reconstruction Consistency with Different Supervision Settings}
In this section, we evaluate the effectiveness of our proposed texture reconstruction consistency under different levels of supervision. The experiments in Tab.~\ref{tab_tex} align with the settings of Ours$_\text{full}$, Ours$_\text{weak}$, Ours$_\text{self}$, and Ours$_\text{self*}$, respectively. In experiments without texture reconstruction, we do not implement 
texture reconstruction consistency losses. Since Ours$_\text{self*}$ solely employs texture reconstruction losses, there is no corresponding experiment without texture.

Under full 3D supervision, the reconstruction of textures does not significantly affect the pose and shape results. This is attributed to the stronger influence of full 3D supervision compared to textures, causing the texture reconstruction less effective.
However, when we reduce the level of supervision to weak 2D supervision, reconstructing textures does not attribute to the pose and shape accuracy. In fact, it introduces noise, resulting in a marginal increase of 0.05 in MPJPE and 0.07 in MPVPE.
Nevertheless, as we further decrease the level of supervision to self-supervision using detected noisy keypoints, the texture reconstruction consistency becomes instrumental in learning accurate pose and shape, which improves results in a reduction of MPJPE by 0.02 and MPVPE by 0.03. This indicates that texture consistency loss improves hand reconstruction in the absence of reliable annotations, reducing the reliance on labeled data. In the last experiment, Self*, we adopt a more rigorous form of self-supervision with solely the self-consistency of textures, without any explicitly defined keypoints as auxiliaries for the training. However, this approach yields unsatisfactory results, with an MPJPE of only 1.66. We believe this is because hand has complex pose configurations and depth ambiguities. Only using texture consistency loss can not learn this information well, which focuses more on pixel-level and perceptual-level features. This also indicates that using only the proposed texture consistency loss terms is insufficient to extract adequate information about hand pose and shape from the images.

\begin{table}
\centering
\caption{Effect of texture reconstruction consistency with different supervision settings on FreiHAND~\cite{zimmermann2019freihand}. The first column indicates the supervision setting, and the second column indicates whether texture reconstruction is included.}\label{tab_tex}
\begin{tabular}{|ll|cc|cccc|}
\hline
Supervision & Texture & MPJPE$\downarrow$ & MPVPE$\downarrow$ & L1$\downarrow$ & PSNR$\uparrow$ & SSIM$\uparrow$ & LPIPS$\downarrow$ \\
\hline
Full & Yes    & 1.21 & 1.23  & 0.03 & 19.40 & 0.94 & 0.11 \\
Full & No    & 1.21 & 1.24  & - & - & - & - \\
\hline
Weak & Yes    & 1.28 & 1.31  & 0.02 & 20.00 & 0.94 & 0.10 \\
Weak & No   & 1.23 & 1.24  & - & - & - & - \\
\hline
Self & Yes   & 1.31 & 1.33 & 0.02 & 20.04 & 0.94 & 0.10 \\ 
Self & No    & 1.33 & 1.35  & - & - & - & - \\
\hline
Self* & Yes    & 1.66 & 1.65  & 0.03 & 17.88 & 0.92 & 0.13 \\
\hline
\end{tabular}
\end{table}

\subsection{Ablation Study}
\subsubsection{Effect of Our Components.} 
We evaluate the effect of our components in Tab.~\ref{tab_ablation}, using the baseline model with full 3D supervision.
The proposed light estimator plays a crucial role in estimating directional lighting and disentangling it from texture, contributing to an increase in PSNR. 
Besides, replacing the EfficientNet-b3 backbone with ResNet50~\cite{He2015resnet} leads to a significant decrease in pose and shape performance.
Furthermore, we explore pretraining the pose and shape estimator on DART~\cite{gao2022dart} instead of RHD~\cite{zb2017RHD}, which leads to a slight increase in performance. DART offers a significantly larger dataset, with more diverse pose and shape distribution than RHD. Here we only use 1/7 of its training dataset for pretraining, so the result highlights its potential for enhancing accuracy. Note that we chose RHD as the pretraining dataset for our baseline to ensure a fair comparison, as it is the most widely utilized pretraining dataset.
We also assess the effectiveness of our losses. 
Removing $\mathcal L_{vert}$ in full supervision results in a slight decrease in MPVPE. Removing $\mathcal L_{perc}$ leads to a noticeable decrease in the quality of the reconstructed texture. Additional ablation studies on losses can be found in the supplementary materials.

\begin{table}[ht]
\centering
\caption{Ablation study on FreiHAND test set~\cite{zimmermann2019freihand}. We study the impact by changing one specific component in each experiment. }\label{tab_ablation}
\begin{tabular}{|l|cc|c|}
\hline
Methods & MPJPE$\downarrow$ & MPVPE$\downarrow$ & PSNR$\uparrow$ \\
\hline
Ours   & 1.21 & 1.23  & 19.547   \\
\hline
w/o light estimator  & 1.22 & 1.25 &  18.983 \\
ResNet50 as backbone  & {1.30} & {1.32}  & 19.582  \\
pretrain on DART  & 1.20  & 1.22  & 19.755 \\
\hline
 w/o $\mathcal L_{vert}$     & 1.21 & 1.25  & -  \\
 w/o $\mathcal L_{perc}$    & 1.21 & 1.23 & 19.512  \\
 \hline

\end{tabular}
\end{table}

\subsubsection{Limitations.}
\begin{figure}
    \includegraphics[width=\textwidth]{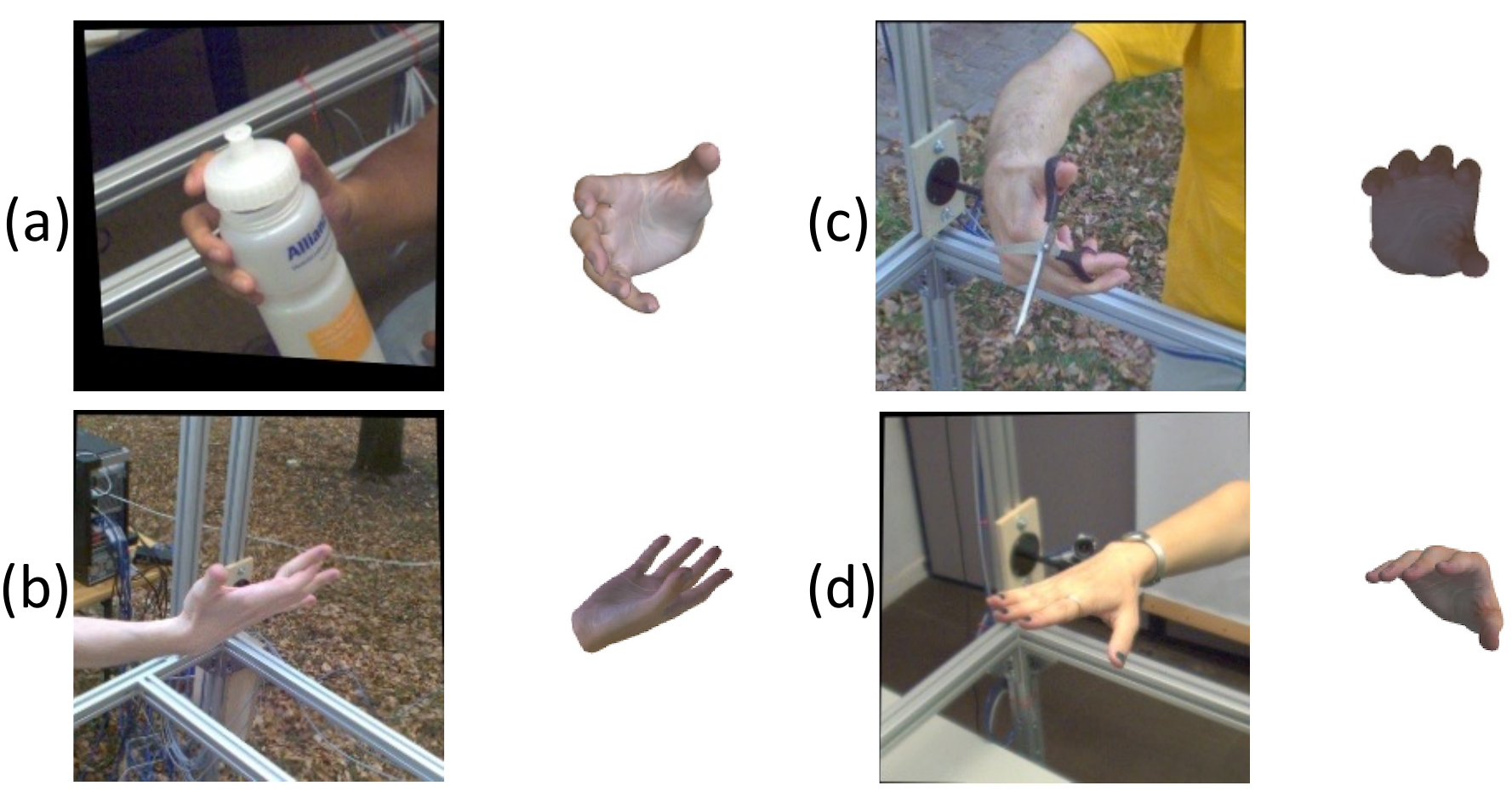}
    \caption{Failure cases. (a) Incorrect texture color due to occlusion. (b) Incorrect pose due to extreme pose. (c) Incorrect pose due to severe hand-object interaction. (d) Unable to reconstruct the black nail.}
    \label{fig_limitation}

\end{figure}

Some limitations and weaknesses remain. First, 
using directional light to represent illumination is limited in some complex scenarios. Furthermore, our method does not consider hand-object interaction scenarios. Fig.~\ref{fig_limitation} shows some failure cases, including object occlusion, extreme pose and texture, and personal features. To address these limitations, we suggest exploring an enhanced pipeline and considering hand-object interaction scenarios.

\section{Conclusions}
We present a 3D hand reconstruction method that is able to generate hands with high-fidelity textures from single input images. Our approach surpasses current state-of-the-art methods in texture quality while maintaining competitive pose and shape accuracy. Through extensive experiments across various levels of supervision, we provide valuable insights into the influence of high-fidelity hand texture reconstruction on pose and shape estimation: in scenarios with noisy and relatively weak supervision, textures effectively contribute to pose and shape learning, whereas their impact becomes less significant under stronger supervision. 
In future work, we aim to extend our research to more complex scenarios that involve hand-object interaction and explore advanced loss terms for texture reconstruction consistency.

%
%
%
%
\bibliographystyle{splncs04}
\bibliography{007-main.bib}

\end{document}